# Exploration of VLM's for Driver Monitoring Systems Applications

Paola Natalia Cañas[1,2*], Marcos Nieto[1], Oihana Otaegui[1], and Igor Rodríguez[2]
1. Fundación Vicomtech, Basque Research and Technology Alliance, Spain, pncanas@vicomtech.org
2. University of the Basque Country (UPV/EHU), Spain

**Abstract**
In recent years, we have witnessed significant progress in emerging deep learning models, particularly Large Language Models (LLMs) and Vision-Language Models (VLMs). These models have demonstrated promising results, indicating a new era of Artificial Intelligence (AI) that surpasses previous methodologies. Their extensive knowledge and zero-shot capabilities suggest a paradigm shift in developing deep learning solutions, moving from data capturing and algorithm training to just writing appropriate prompts. While the application of these technologies has been explored across various industries, including automotive, there is a notable gap in the scientific literature regarding their use in Driver Monitoring Systems (DMS). This paper presents our initial approach to implementing VLMs in this domain, utilising the Driver Monitoring Dataset to evaluate their performance and discussing their advantages and challenges when implemented in real-world scenarios.

**Introduction**

In the automotive industry, Advanced Driver Assistance Systems (ADAS) and Automated Driving (AD) are mostly based on exterior perception algorithms that enable high levels of vehicle automation (e.g., SAE L2+/L3/L4 [1]). These algorithms make the vehicle able to determine its trajectory within the road and lane, recognize traffic signs, and detect other road participants (like pedestrians, bikes, cars, trams, etc.). Equally important are in-cabin perception algorithms that constitute a passenger/driver monitoring system (DMS). These systems are essential for the vehicle to be aware of the state of its occupants, implementing functionalities such as driver distraction detection, drowsiness detection, gaze estimation, and occupancy estimation. Driver monitoring systems have gained relevance since they are becoming mandatory to prevent accidents due to human errors and, in autonomous driving scenarios, to assess driver readiness and fitness to regain control of the vehicle in case of need.

Traditional DMS rely on a combination of sensors and algorithms to detect signs of drowsiness, fatigue, distraction, intoxication, and other potentially dangerous conditions. However, these systems often face limitations in terms of adaptability, and the ability to handle diverse real-world scenarios. They often require calibration processes, depend heavily on the type and position of sensors, and typically function as classifiers that can identify a predefined set of activities. But humans can behave in unexpected ways. For example, distraction detection might focus on activities like texting or eating, etc. While this could certainly be a robust system, and help prevent many accidents, it falls short in accounting for the wide variety of activities humans might engage in that pose risks, or diverse situations that can affect driving which the system might not consider. Also, as vehicles become more autonomous, drivers are allowed to engage in more activities unrelated to driving, expanding the list of possible distractions. Similarly, in intoxication detection, human behaviour under the influence can manifest in a variety of symptoms or signals, many of which might be out-of-distribution for the system tasked with detection. Current algorithms may not fully capture the breadth of intoxication indicators, leading to potential false negatives identifying dangerous conditions.

Visual Language Models have gained popularity in 2020–21 with the emergence of models like OpenAI's CLIP [2]. These models can process images (vision) and natural language text (language), performing tasks such as image captioning or image creation from text, among others [3]. Combining visual and language models into one architecture can help the models understand the context, semantic representations, and more abstract concepts of our world; this logic is inherent in the understanding of human language.

VLMs have the potential to revolutionize driver and in-cabin monitoring by offering a more holistic understanding of the driving scene. Rather than focusing on individual variables, VLMs are trained to describe the entire scene, considering all crucial elements. This comprehensive approach allows them to construct a coherent narrative around the scene, leading to a more thorough assessment of the driver's situation.

Despite the potential benefits, there is a notable lack of scientific research exploring the application of VLMs in this field. We aim to conduct an initial exploration of how these systems perform in tasks such as distraction detection, drowsiness detection, and gaze estimation. By evaluating their performance, we hope to determine whether they can match or even surpass state-of-the-art models, or identify areas where they fall short. To achieve this, we will utilize data from the Driver Monitoring Dataset (DMD), which contains extensive material of drivers in various states of drowsiness and distraction containing drivers doing several actions that imply distraction like texting, having a phone call, drinking water, besides driving safely, as well as detailed gaze annotations. By integrating VLMs into DMS, we expect the model to:
- Have better scene comprehension, enabling it to provide detailed descriptions and respond to queries through Visual Question Answering (VQA) tasks.
- Simultaneously classify multiple variables.
- Address scenarios that might be out-of-distribution from the training dataset, thanks to their extensive knowledge about the world and humans.
- Reason about and explain their predictions, contributing to the system's overall explainability.

**Methodology**

To evaluate the performance of VLMs in driver monitoring systems applications, we employed various techniques including zero-shot prompting, few-shot prompting, and prompt engineering. None of these techniques require training, meaning that no weights are updated.
- **Zero-shot prompting:** it is asking the model to make predictions or perform tasks it has not explicitly trained for or has seen in its training process. This is possible since these models had an extensive training on large amounts of datasets, allowing them to generalize knowledge to novel scenarios without needing specific task examples. We assume that driver monitoring tasks were not part of these VLMs models' training.
- **Few-shot prompting:** If zero-shot prompting does not yield satisfactory results, few-shot prompting can be employed. This strategy involves providing the model with a limited number of examples (usually 1 or 2) for a specific task. It is similar to teaching the model how the response should look using a few samples and then allowing it to predict new data. This approach helps the model refine its responses for specific tasks while utilizing its pre-existing knowledge.
- **Prompt engineering:** this consists of designing input prompts to effectively produce the desired responses from a model. Finding the best prompt, usually composed of text and sometimes images, is crucial for getting the full potential of VLMs. Preparing good prompts can guide the model to generate accurate outputs. A clear explanation of the task the researchers want the model to perform is very relevant, also specifications about the output format can be given.

One critical aspect to consider in this research is the selection of the model. Many commercial models, such as ChatGPT and Gemini, are private and can be accessed through their respective websites or via subscription-based APIs, like the OpenAI API for ChatGPT [4]. However, there are alternative options to explore. The scientific community continues to make progress, building an extensive pool of VLMs with varying sizes, purposes, and knowledge bases.

In this investigation, we opted to use smaller models to ensure computational efficiency, making them more suitable for implementation in vehicle computational setups. Larger models, while potentially more powerful, require substantial computational resources, which are not ideal for in-vehicle applications. The model selected for this study is Idefics2 [5], and is available on the Hugging Face platform, with approximately 7 billion parameters.

Exploration of VLM's for Driver Monitoring Systems Applications

*Experiments*

We collected test data from the DMD. We iteratively make predictions over all the images of a video and subjectively evaluate the performance. These are the experiments:

- **Zero-shot:**

The gaze annotations in the DMD consist of the regions inside the car where the driver is looking (e.g., left mirror, steering wheel, etc.). We tested the models to predict which zone of the car the driver was focusing on. Despite multiple attempts and the implementation of best practices in prompt engineering, we did not achieve consistently accurate results for this specific task of gaze estimation by zones using the images provided by the DMD. This image could not give enough information to the model to estimate a gaze region. Therefore, we tried with a simpler task of just indicating the direction of the gaze. This was done with the prompt: *"Where is the person looking to in this image?"*.

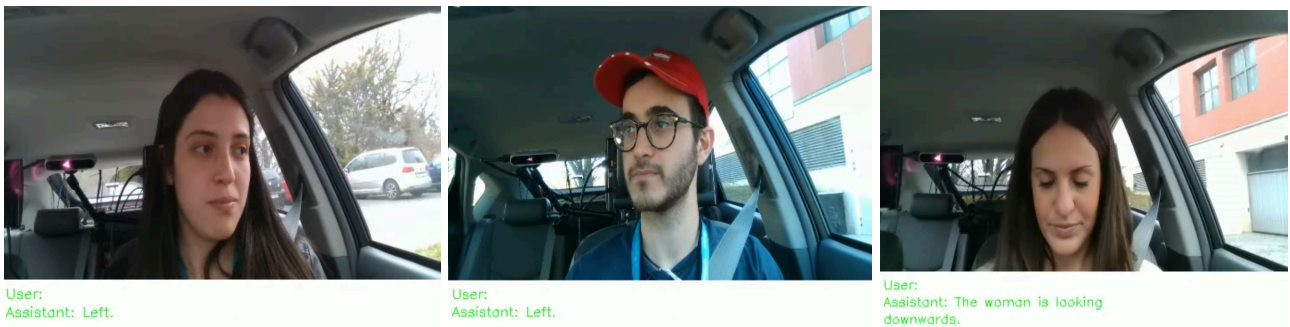

**Figure 1 - Inference with zero-shot. Response: (Left) Left. (Center) Left. (Right) The woman is looking downwards**

Figure 1 presents the results. For this simpler task, the model's performance was inconsistent as well. The model predicted two images both as left, where the drivers are looking to opposite sides. This inconsistency makes it difficult to assess which prediction is correct, as the reference could be either the driver's left or the image's left side. However, the model found it easier to correctly identify when the driver was looking downward. Then, we tested the model's ability to detect distraction.

- **One-shot:**

One-shot prompting is done by providing 1 example to the model with a chat-like template simulating a correct response from the model. This chat gives the instruction (user), the correct response (assistant) and the following instruction (user). The prompt includes the image from the example and the image for inference. In this case, the example given to the model was an image of a driver talking on the phone.
The complete prompt is shown in Figure 2:

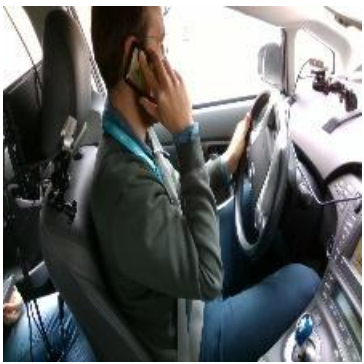

*-User: You are a driver monitoring system that is responsible for assuring the driver is driving safely and alert when they are distracted. What is the state of this driver? <example_image>*
*-Assistant: This driver is distracted because he is having a phonecall while driving*
*-User: And how about this driver? <inference_image>*

**Figure 2 - Prompt for the one-shot experiment. On the left, there is the sample image**

The inference image is shown in Figure 3, along with the response of the model: "*The driver is distracted because he is adjusting his hair while driving*". This is correct.

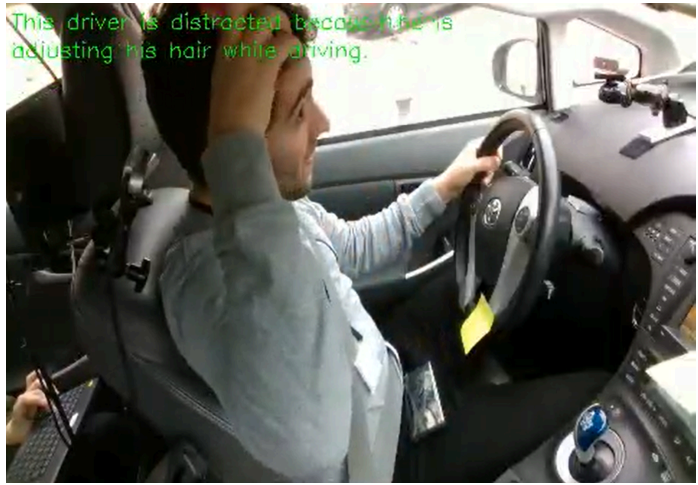

**Figure 3 - Inference of model with one-shot prompting. Response: The driver is distracted because he is adjusting his hair while driving**

We also notice that in other cases the response was conditioned by the example provided. When predicting with the inference image shown in Figure 4, its response was: "*The driver is alert because he is not using his phone while driving*". This is not incorrect, the driver is not distracted, but for solutions that require an explanation of the actual situation, this response is not helpful.

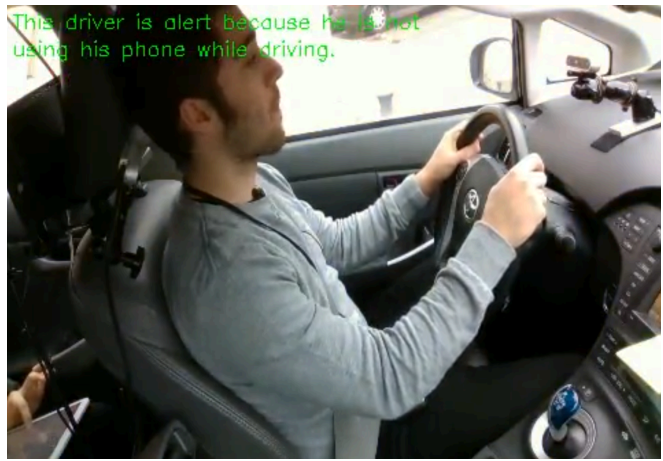

**Figure 4 - Inference of model with one-shot prompting. Response: The driver is alert because he is not using his phone while driving**

Motivated by the interesting results so far, we wanted to also test with an inference image from a different camera position. This image and the response are shown in Figure 5, this was: "*The driver is not distracted because he is not using his phone*". The model could also perform well with an image with a different perspective from the one from the example.

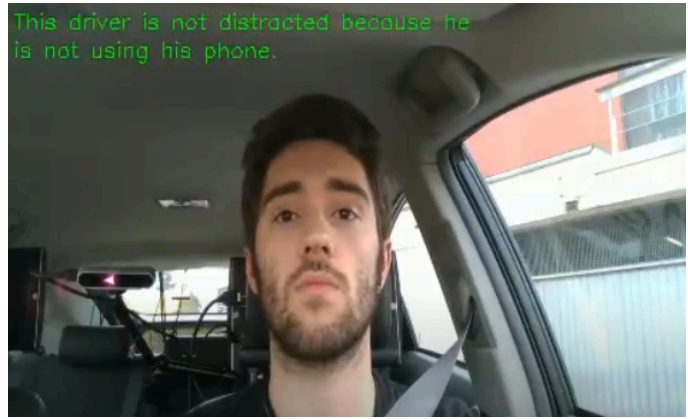

**Figure 5 - Inference of model with one-shot prompting. Response: The driver is alert because he is not using his phone while driving**

Using a video from the drowsiness material of the DMD, we also tested this prompt. Even though the model was instructed to identify distraction and the example given is a distraction activity, the model is also capable of also alert when a dangerous situation is given by a sign of drowsiness as yawning. This can be seen in Figure 6, the response was: "*The driver is drowsy because he is yawning while driving*".

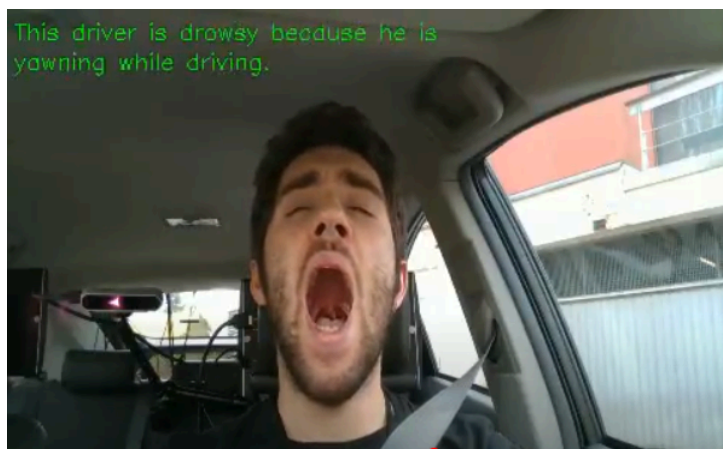

**Figure 6 - Inference of model with one-shot prompting with drowsiness video. Response: The driver is drowsy because he is yawning while driving**

- **One-shot + output code-format instruction**:

The nature of VLMs, which provide text sentences as output, is beneficial for explainability and allows for richer descriptions. However, current DMS require simpler responses, such as a Boolean (e.g., true or false for distraction). We aimed to test whether this formatting of the response was also feasible. For this experiment, the prompt included instructions for the model to give the response in code format, setting variables to true or false. We also employed a one-shot approach, providing the model with an example. In this instance, an image of a driver talking on the phone was used as the example. In Figure 7 is the prompt with the formatting instructions:

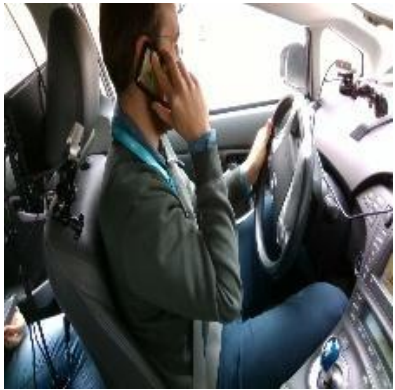

*-User: You are a driver monitoring system that is responsible for assuring the driver is driving safely and alert when they are distracted. You need to communicate with the HMI to alert the driver, please provide the following variables with True or False: Distracted, Talking, Using phone. What is the state of this driver? <example_image>*
*-Assistant: Distracted = True, Talking = True, Using phone=True*
*-User: And how about this driver? <inference_image>*

**Figure 7 - Prompt for the one-shot experiment. On the left, there is the sample image**

The results can be seen in Figure 8. With an image similar as the one as the example, the model performs well predicting: "*Distracted=True, Talking=True, Using_phone=True*". Also, with an image when the driver is safely driving, with: *"Distracted=False, Talking=False, Using_phone=False"*.

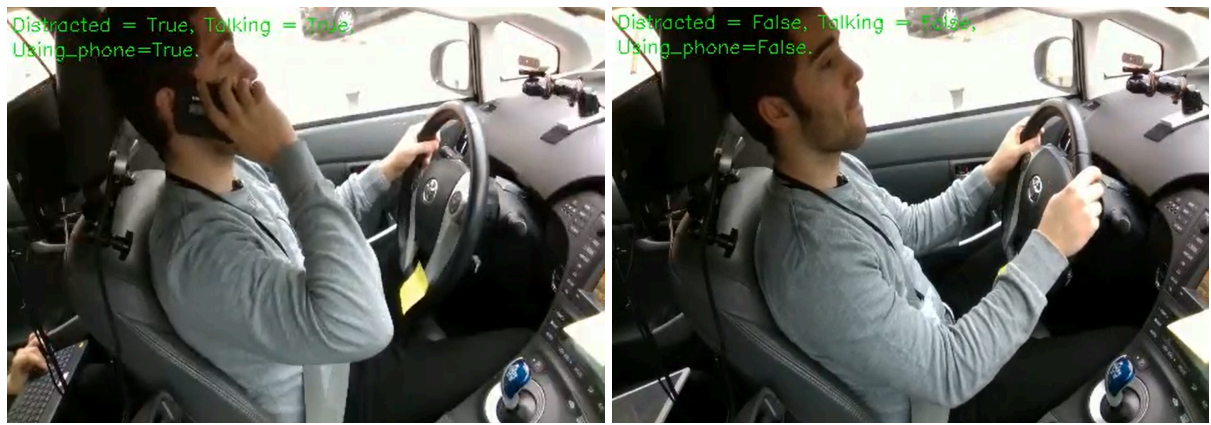

**Figure 8 - Example of inference with a code-formatted answer with one-shot learning. Response (left): Distracted=True, Talking=True, Using_phone=True. Response (right): Distracted=False, Talking=False, Using_phone=False**

Although the predictions shown for this experiment are correct, we did not see consistent accurate predictions across all the tested images. The "Distracted" variable was correctly predicted most of the time, but the variables "Talking" and "Using Phone" posed more challenges. The performance could potentially improve if the model is prompted to respond to only one variable at a time.

**Challenges & Opportunities**

After this exploration and with our experience generating data and algorithms for DMS's we found the following challenges and opportunities:

*Lack of definitions/standards*

One of the primary challenges in developing effective VLMs for DMS is the lack of standardized definitions for key concepts such as distraction. Without clear and consistent definitions, it becomes difficult to create precise instructions for VLMs, leading to variability and inconsistency in predictions. For instance, an activity like adjusting one's hair might be considered a distraction by some systems but not by others, as it is not an illegal activity while driving. This lack of standardization complicates the model's ability to accurately identify and respond to distractions that are not specifically explicit in the training set or instructions.

*Multi-agents architectures*

We have observed that models often struggle to predict multiple variables simultaneously with high accuracy. Predicting multiple variables such as distraction, drowsiness, gaze direction, or other aspects like gender and the number of passengers, all at once, can be challenging.

Implementing a multi-agent architecture could address this issue. In this approach, multiple specialized models each focus on a specific aspect of driver behavior, a particular variable, or a different task. By narrowing the focus of each model, the complexity of predictions is reduced, and the precision of each model is increased. This strategy, illustrated in Figure 9, involves a set of models that describe the situation in one aspect, with another model that integrates this information to make a conclusion or risk assessment.

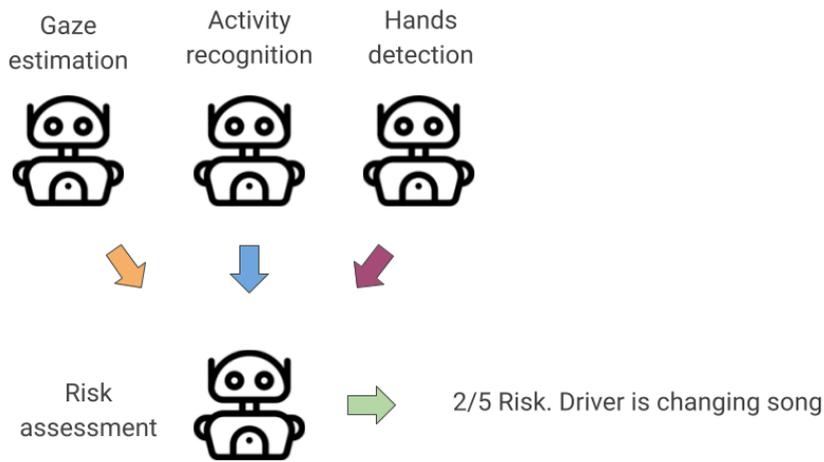

**Figure 9 - Example of Multi-agents architectures**

*Evaluation*

Continuing with the need for a discrete or a simpler response from the model, this can be also relevant for the evaluation of the model. Not going further, to automatically evaluate these responses we need to compare them with the labels. The current approach faces difficulties due to the lack of direct comparability between model predictions and labels, complicating the evaluation process. A quick and potential solution is to include another LLM to evaluate the responses, determining whether they correspond accurately to the labels. Figure 10 illustrates this process. The instructions for this LLM can be: *You are an agent in a multi-agent system. You will receive the description of the state of a driver from the driver monitoring agent, and you will need to say if that description can be classified correctly as a label that will be provided by another agent. Please only say 'True' for positive and 'False' for negative.* This way is possible to calculate better performance metrics.

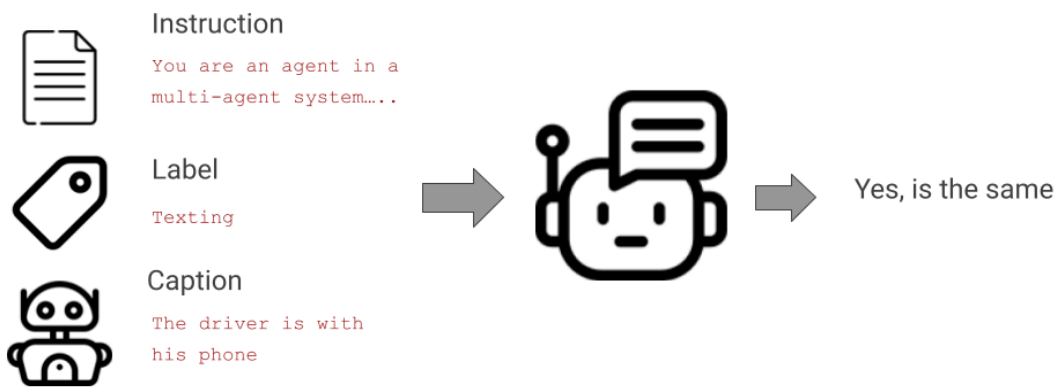

**Figure 10 - Example of evaluation strategy using an LLM**

*Data*

In general, there is a lack of public and quality data for DMS. In addition to that, and related to the prior challenge, data specifically tailored for VLMs for DMS is needed for developing effective systems. Current datasets are predominantly designed for supervised learning and typically include only single-label annotations, like the DMD. However, VLMs operate using images and text in the form of descriptions, necessitating richer and more detailed contextual information for training and evaluation.

The inherent multimodality of VLMs presents a unique opportunity to expand the scope of DMS by incorporating diverse data sources. This includes integrating various types of data, such as images, text, and possibly even sensor data, to provide a more comprehensive understanding of the driving environment. Collecting and utilizing more comprehensive datasets that offer detailed descriptions, and contextual information will significantly enhance model performance and open new avenues for investigation and application.

*Performance*

The current performance of VLMs is time-consuming, making them unsuitable for real-time applications like gaze estimation. Additionally, the large size of these models poses a challenge for in-vehicle computation resources, which often have limited processing capabilities. However, with some optimism for the future, hardware improvements and software optimizations are expected to help overcome these barriers, potentially enabling these models to run on lower-performance resources soon.

Another viable solution is to consider commercial services that offer significantly reduced inference times. These include private models like ChatGPT or custom-trained models with a sub-100ms latency [6]. Such services can enable VLMs to function effectively in driving applications, providing timely and reliable predictions.

**Conclusions**

In this technical paper, we conducted an exploratory exercise from which we can conclude that VLMs show significant potential in building DMS's, although a deeper evaluation is required to obtain a comprehensive performance overview. Despite VLM's current inability to provide stable predictions and readiness for immediate implementation, with proper training and improvement, VLMs can not only reach but surpass the current state-of-the-art, offering several benefits:
- **Camera-position invariant predictions:** VLMs are capable of making accurate predictions regardless of camera position, enhancing the robustness of DMS.
- **Handling out-of-distribution situations:** VLMs can consider real-world scenarios that are out-of-distribution, improving the system's adaptability to various driving conditions.
- **Multimodality:** VLMs are already equipped for multimodal tasks, opening new avenues for exploration in DMS, which have not yet been thoroughly investigated.
- **Enhanced Explainability:** VLMs have the potential to provide more transparent and understandable predictions by reasoning about and explaining their outputs. This contributes to the overall explainability of the system, making it easier for developers and users to trust and interpret the model's decisions.
- **Slow performance compensation:** There are strategies to compensate for slower performance in real-time applications, whether through awaiting hardware improvements or utilizing third-party inference services. This will enable VLMs to function effectively in driving application contexts.

**Acknowledgements**

This work was funded by the Horizon Europe programme of the European Union, under grant agreement 101076868 (project AWARE2ALL).
Funded by the European Union. Views and opinions expressed here are however those of the author(s) only

**References**

1. SAE International. (2021). Taxonomy and Definitions for Terms Related to Driving Automation Systems for On-Road Motor Vehicles (SAE Standard J3016_202104). https://doi.org/10.4271/J3016_202104

2. Radford, A., Kim, J. W., Hallacy, C., Ramesh, A., Goh, G., Agarwal, S., Sastry, G., Askell, A., Mishkin, P., Clark, J., Krueger, G., & Sutskever, I. (2021). Learning Transferable Visual Models From Natural Language Supervision. Proceedings of the 38th International Conference on Machine Learning, PMLR. https://arxiv.org/abs/2103.00020

3. Alayrac, J.-B., Donahue, J., Luc, P., Miech, A., Barr, I., Hasson, Y., Lenc, K., Mensch, A., Millican, K., Reynolds, M., Ring, R., Rutherford, E., Cabi, S., Han, T., Gong, Z., Samangooei, S., Monteiro, M., Menick, J., Borgeaud, S., Brock, A., Nematzadeh, A., Sharifzadeh, S., Binkowski, M., Barreira, R., Vinyals, O., Zisserman, A., & Simonyan, K. (2022). Flamingo: a Visual Language Model for Few-Shot Learning. arXiv preprint arXiv:2204.14198

4. OpenAI. (2025). ChatGPT (January 2025 Version). OpenAI.

5. Laurençon, H., Marafioti, A., Sanh, V., & Tronchon, L. (2023). *Building and better understanding vision-language models: insights and future directions*.

6. Together AI. (n.d.). Together AI – The AI Acceleration Cloud. Retrieved from https://www.together.ai